\documentclass[letterpaper, 10 pt, conference]{ieeeconf}  
\IEEEoverridecommandlockouts                               
\usepackage{graphicx}                       
\usepackage{epsfig}                         
\usepackage[tight,footnotesize]{subfigure}  
\usepackage{amssymb,amsmath}
\usepackage{mdwmath}
\usepackage{commath}                        
\usepackage{eqparbox}
\usepackage{amsmath,amsfonts,amssymb}

\newcommand{\et}{{\em et al.\ }}

\newcommand{\beq}{\begin{equation}}
\newcommand{\eeq}{\end{equation}}
\newcommand{\bear}{\begin{eqnarray}}
\newcommand{\bears}{\begin{eqnarray*}}
\newcommand{\eear}{\end{eqnarray}}
\newcommand{\eears}{\end{eqnarray*}}
\newcommand{\bdm}{\begin{displaymath}}
\newcommand{\edm}{\end{displaymath}}
\newcommand{\lba}{\left[\begin{array}}
\newcommand{\ear}{\end{array}\right]}

\usepackage{bm}
\usepackage{stfloats}                       
\usepackage{hyperref}
\usepackage{cite}                           
\usepackage[T1]{fontenc} 
\usepackage{tabularx}
\usepackage[table]{xcolor}
\usepackage{longtable}
\usepackage{booktabs}       					
\usepackage{xcolor,colortbl}               		
\newcommand{\red}{\textcolor[rgb]{ .773,  0,  .043}}

\title{\LARGE \bf Learning Human-Robot Collaboration Insights through the Integration of Muscle Activity in Interaction Motion Models}
\author{Longxin Chen, Juan Rojas, Shuangda Duan, and Yisheng Guan. \\
\thanks{All authors are with the School of Electromechanical Engineering in Guangdong University of Technology in Guangzhou, China.}%
}
\begin{document}
\maketitle
\thispagestyle{empty}
\pagestyle{empty}
\begin{abstract}
Recent progress in human-robot collaboration (HRC) makes fast and fluid interactions possible, even when human observations are partial and occluded.
Methods like Interaction Probabilistic Movement Primitives (ProMPs) model human Cartesian trajectories through motion capture systems. However, such representation does not properly model tasks where similar motions are used to handle different objects. As such, under current approaches, a robot would not be able to properly adapt its pose and dynamics for proper handling.
We propose to integrate the use of Electromyography (EMG) into the Interaction ProMP framework and utilize EMG-based muscular signals to augment the human observation representation. The contribution of our paper is the increased capacity to discern tasks that have similar trajectories but ones in which different tools are utilized and require the robot to adjust its pose for proper handling. 
Multidimensional Interaction ProMPs are used with an augmented vector that integrates muscle activity. Augmented time-normalized trajectories are used in training to learn correlation parameters and robot motions are predicted by finding a best weight combination and temporal scaling for a task. 
Collaborative single task scenarios with similar motions but different objects were used and compared. For one experiment only joint angles were recorded, for the other EMG signals were additionally integrated. Task recognition was computed for both tasks. Observation state vectors with augmented EMG signals were able to completely identify differences across tasks, while the baseline method failed every time.
Integrating EMG signals into collaborative tasks significantly increases the ability of the system to recognize nuances in the tasks that are otherwise imperceptible, up to 74.6\% in our studies. Furthermore, the integration of EMG signals for collaboration also opens the door to a wide class of human-robot physical interactions based on haptic communication that have been largely unexploited in the field.
Supplemental information including video, code, and results analysis can be found at \cite{2017Humanoids-Chen-supplementalURL}.
\end{abstract}
\section{INTRODUCTION}
Interest in HRC has significantly increased in recent years. The promise of synergistically combining the best of what robots and humans have to offer has led to numerous studies. However, many challenges remain in facilitating programming robot collaborative partners. The variety of tasks in which a human needs assistance is practically unlimited. Robots must easily learn and adapt to unstructured scenarios. Recent progress in HRC now makes fast and fluid interactions possible, even when human observations are partial and occluded. 

Numerous approaches used to generate robot motion in response to human motion observations have relied on the use of joint angle or Cartesian trajectory information. Works like Dynamic Movement Primitives (DMP) \cite{Amor2016Interaction,Schaal2006Dynamic,Ijspeert2013Dynamical}, Interactive Meshes \cite{Vogt2015Behavior,Vogt2017A,Vogt2017Learning}, and Interaction ProMPs \cite{2017IJRR-Maeda-PhaseEstimation,2015ICRA-ewerton-LearnMultCollabTasks_MixtureInteractionPrimitives,Maeda2016Probabilistic} use motion capture systems to record human motion trajectories. However these systems are unable to properly model tasks where similar motions are used to perform different tasks, such as that of passing, holding, or coordinating motion of a human using different tools with different shape and inertial properties. As such, under current approaches, a robot is unable to properly adapt its pose and dynamics when two tasks with similar motion but different objects are used.

\begin{figure}[t!]
\centering
	\includegraphics[width=8.2cm]{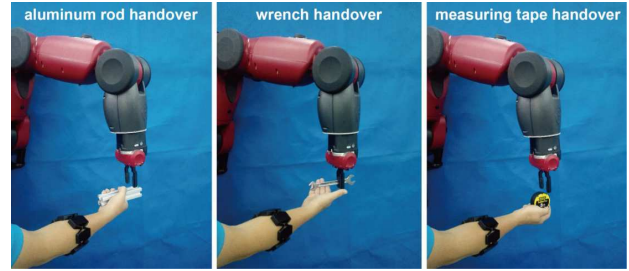}
 	\caption{A robot collaborator is empowered when it is able to discern different tasks that consist of similar human trajectories. In this figure, three tasks are shown where a human uses a similar trajectory to hand over three distinct objects to the robot. In each case, we augmented the observation vector with EMG-based muscular activity signals that enabled to robot to discern across tasks and choose the correct robot response.}
	\label{fig:collab_task_example}
\end{figure}

In this paper we explore techniques that enable increased task recognition discernment given human observations. Particularly, we explore the impact of integrating EMG-based muscular activity signals when used alongside motion trajectories in the Interaction ProMP framework. The contribution of this paper is the discernment of tasks that have similar motion trajectories, but ones in which objects of different shapes and dynamics are used. Better action recognition also leads to more natural interactions as a robot can adjust its pose and dynamics to minimize (mental, emotional, and physical) load placed on the human to compensate for poor adjustment on the robot's part. Fig. \ref{fig:collab_task_example} illustrates a hand-over interaction in a collaborative task. 

Multidimensional Interaction ProMPs are used with an augmented state vector that integrates EMG-based muscle activity. This works builds on the Phase Estimation approach of \cite{2017IJRR-Maeda-PhaseEstimation}. Provided a set of human-robot collaborative task demonstrations, time aligned trajectory way-points and EMG signals are parameterized into a lower dimensional weight space as a linear combination of basis functions. A Gaussian distribution is built from the set of weight vectors obtained in training and a normal distribution is also built from time-scaling values used to normalize training data yielding a probabilistic movement primitive. As for the Interaction segment, the correlation of all human-robot data dimensions is computed and the robot motion is inferred by computing a posterior probability distribution over the weights conditioned on the partial augmented human observation. The weight distribution requires a new mean and covariance from the partial observations, both of which are computed by using a Kalman filter. For task recognition, the task with the highest posterior probability for new observations given the task probability is selected. In Maeda \et's work, temporal variance is included in the model. A phase ratio needs to be computed from the sparse sequence of observations, to determine an associated observation matrix to finally condition and do prediction. 

To test the effects of EMG signals in Interaction ProMPs, three distinct hand-over tasks were performed, all of which consisted of similar motions but used different objects. Experiments were done with and without EMG-signals. Task recognition was reported for both scenarios for different number of demonstrations and observation ratios. 
Integrating EMG signals into collaborative tasks significantly increases the ability of the system to recognize nuances in the tasks that were otherwise imperceptible, up to 74.6\%. EMG signals recognized tasks better in 11/12 of our comparative studies and did it overwhelmingly better. We also purport that user-loads (mental, physical, and emotional) would diminish significantly as humans would not need to adjust their handling to make up for the robot's deficiency.  Finally, the integration of EMG signals in HRC opens the door to a wide class of human-robot physical interactions based on haptic communication that have been largely unexplored in the field.
Supplemental information including video, code, and results analysis can be found at \cite{2017Humanoids-Chen-supplementalURL}.

\section{Related Work}
HRC poses a dual problem: one of action recognition and movement generalization. This section we discuss the conventional interaction motion models and previous works related to the use of EMG signals in HRC. 

In \cite{Amor2016Interaction,Schaal2006Dynamic,Ijspeert2013Dynamical}, DMPs are introduced as a time-dependent movement representation. DMPs comprise a proportional-derivative (PD) controller and a non-linear forcing function. Based on the DMP framework, Interaction Primitives (IPs) capture the variance of DMP parameters and generate a probability distribution. The probabilistic model learns the inter-agent correlations and allows to generalize skills in HRC. 

In \cite{Vogt2017Learning}, Interaction meshes (IM) were used to learn human-robot interactions from human-human demonstrations. IMs capture spatio-temporal relationships between the body movements of two interacting partners. For any given time-step, an IM represents a pair of postures in the human-human demonstration. IMs allow to transfer a collaborative skill from one pair of partners to another (i.e. a human-robot pair) given the set of IMs. IMs are coupled with Hidden Markov Models (HMMs) to have both the ability to generate robot motions (through the IMs) and perform task recognition (through the HMMs). HMMs have been a popular modeling approach in which the process is assumed to be Markov and consist of unobserved hidden states that are inferred \cite{Calinon2004Gesture,calinon2006learning}. Furthermore, IMs can be deformed to adapt to varying trajectory observations in the interacting partners \cite{Vogt2015Behavior,Vogt2017A,Ho2010Spatial}.

In \cite{Paraschos2013Probabilistic,Koert2017Demonstration}, ProMPs were introduced as an alternative to DMPs. ProMPs are a time-dependent movement representation that do not need a forcing function, instead trajectories are approximated by a weighted sum of time-dependent basis functions. 
More recently, Maeda \et proposed Interaction ProMPs based on ProMPs for HRC \cite{2017IJRR-Maeda-PhaseEstimation,2015ICRA-ewerton-LearnMultCollabTasks_MixtureInteractionPrimitives,Maeda2016Probabilistic}. Interaction ProMPs capture temporal and spatial variances of motion trajectories as well as correlations across all human and robot dimensions. The model can recognize executed tasks and generate corresponding robot motion given human motion observations. That is, both motion generalization and action recognition are jointly implemented in the framework.

All previous works are limited in that they only modeling motion trajectories. In situations where different tasks are executed with similar trajectories, these techniques are unable to discern across tasks. This is important given that in collaboration, it is not uncommon to to perform similar motions with different tools. Consider any kind of hand-over task, the same motion is used for a variety of tools that have unique shape and dynamical properties. Thus, it is of significant interest to explore techniques that enable greater insight into tasks with similar spatio-temporal relationships in motion trajectories. 
In our work, we propose to integrate the use of EMG-based muscular activity in the previously presented Interaction ProMP model. By integrating EMG signals, the system is able to gain insights unavailable in spatio-temporal trajectory patterns in motion trajectories. Muscular activity contains signatures that differentiate both pose and dynamical patterns hence providing key information to our model.

We note that there seem to be no other works in which EMG-signals are used to model and classify human motions within HRC. 
Some studies like that of  Reed \et \cite{2008TroHap-Reed-PhysicalCollab-HH_HR}, measured human force profiles in human-human interactions, where humans developed a specialization of roles. Later when a human-interacted with robots, no specialization took place according to the force profiles. This is an example where human force feedback was used, but not to affect the response of the collaborative robot. 
Peterne \et \cite{Peternel2016Adaptation}, used EMG signals to estimate human partner fatigue in human-human collaborative tasks. 
Kulic \et \cite{2007Tro-kulic-AffectiveStateEstimationHRI}, use human physiological signals like heart rate, perspiration rate, and facial muscle contract to measure the body-language interaction between a human and a robot. A robot manipulator was conditioned to move to different distances from the human, and the physiological response was measured. This study is similar to our current work in that human signals are modeled, but differ in the this study did not use them to tell the robot how to move. Instead the goal was simply to model the affective state of the human given a robot motion. 

\section{Methodology}
In HRC tasks, Interaction ProMPs generate a robot collaborative motion based on the prediction from a set of partial human movement observations. The approach also works in multi-task scenarios. Our work explains the steps need to integrate and process EMG-based  muscle activities in addition to motion trajectory data. 
\subsection{Probabilistic Movement Primitives for a Single Dimension}\label{subsec:ProMP_1dim}
ProMPs summarize patterns across demonstrations in a probabilistic manner. They are able to capture correlations across all data dimensions and describe variations in which movements can be executed leading to a probability distribution over trajectories. Representing variance information correctly is critical as it reflects the importance of single time steps for a movement execution. For each time step, a single dimensional position is represented by $y_t \in \mathbb{R}^1$ and a trajectory of $T$ time steps as $\bm{y}_{1:T}$. We adopt linear regression with $n$ Gaussian basis functions $\bm{\psi}$ to represent one motion trajectory. The probability of observing a trajectory $\bm{y}_{1:T}$ given an underlying weight vector $\bm{\omega}$ is given as a linear basis function model:
\begin{equation}
  \begin{aligned}
  y_t &= \bm{\psi}_t^T \bm{\omega} + \epsilon_y,\\
  p( \bm{y}_{1:T} | \bm{\omega}) &= \prod\limits_1^T \mathcal{N}(y_t | \bm{\psi_t^T} \bm{\omega}, \sigma_y),
  \end{aligned}
\end{equation}
where, $\epsilon_y \sim \mathcal{N}(0, \sigma_y)$ models zero-mean i.i.d. Gaussian noise. The set $\bm{\psi} = {[{(\psi_t)}_1, {(\psi_t)}_2,...,{(\psi_t)}_N]}^T \in {\mathbb{R}}^{N \times 1}$  contains values of each of the basis function at time $t$. Given a basis function, one can compute $\bm{\omega}$ for each trajectory $\bm{y}_{1:T}$ using linear regression as:
\begin{equation}
\bm{\omega} = ({\bm{\Psi}_{1:T}^T \bm{\Psi}_{1:T})^{-1}} \bm{\Psi}_{1:T} \bm{y}_{1:T},
\end{equation}
where,
\begin{equation}
\bm{\Psi}_{1:T}
=\begin{bmatrix}
{(\psi_1)}_1 & \cdots  &  {(\psi_1)}_N\\
  \vdots     & \ddots  &  \vdots  \\
{(\psi_T)}_1 & \cdots\ &  {(\psi_T)}_N\\
\end{bmatrix}
\end{equation}
The $\bm{\omega}$ vector can compactly represent a single trajectory. Having a set of motion trajectories, we can compute a probability distribution over the weights $\bm{\omega}$. To capture the variance across trajectories in different demonstrations, we define $\bm{\theta}$ as a parameter that governs the distribution of weight vectors in the set $\bm{\omega}$ and we assume that $\bm{\omega} \sim \mathcal{N}({\bm{\mu}_\omega, \bm{\Sigma}_\omega})$, that is $\bm{\theta} = (\bm{\mu}_\omega, \bm{\Sigma}_\omega)$. 

The trajectory distribution $p(\bm{y}_{1:T}; \bm{\theta})$ can now be computed by marginalizing out the weight vector $\bm{\omega}$. The distribution $p(\bm{y}_{1:T}; \bm{\theta})$ defines a Hierarchical Bayesian Model (HBM) whose parameters are given by the observation noise variance $\sigma_y$ and the parameters $\bm{\theta}$ of $p(\bm{\omega}; \bm{\theta})$. For now, we can compute the probability distribution of a position at a given time from the distribution of $\bm{\omega}$ as 
\begin{equation}
	\begin{aligned}
		p(y_t | \bm{\theta}) &= \int p(y_t | \bm{\omega}) p(\bm{\omega} | \bm{\theta}) d \bm{\omega}\\
		&= \mathcal{N}(y_t | \bm{\psi}_t^T \bm{\mu}_\omega, \bm{\psi}_t^T \bm{\Sigma}_\omega \bm{\psi}_t + \sigma_y).
	\end{aligned}
\end{equation}
The above framework captures spatial correlations from a set of demonstrations. To cope with demonstrations of different durations, the training set must be time aligned (done through resampling in this work).
\subsection{Correlating Muscular Activity into Interaction Motion Model}\label{subsec:emg_in_promp}
In this section, we extend ProMPs to a multidimensional setting and compute the correlation for the full set of data-dimensions for human and robot across demonstrations. Previous works assume that human-motion collaborative-task trajectories differ spatio-temporally from one another. Under this assumption, the use of Cartesian information from human motion capture systems has been sufficient to distinguish different tasks. However, if the assumption is violated and different tasks share similar trajectories, the task recognition system is bound to fail. We consider the introduction of EMG-based muscular activities as part of the observed state in Interaction ProMPs. EMG signals are easily integrated as a temporal sequence. With them, we attempt to infer future robot responses from human observations (now Cartesian pose and EMG information),  with more nuanced insights into the collaborative task. Namely, the ability to discern different tasks with similar pose observations but with distinct muscular activities.

Now we introduce the mathematical model for Interaction ProMPs with the augmented EMG-signals. For human observations, consider $p$ pose dimensions and $e$ EMG signal channels, while for robot observations, consider $j$ joint angles. Each collaborative demonstration consists of $(p+e+j)$ dimensions in the training trajectories. For HRC, the state vector $\bm{y_t}$ at time $t$ is the concatenation of the $(p+e)$  human observations and the $j$ joints of the robots, such that
\begin{equation}
\bm{y}_t = {[y_{1,t}^H, ... y_{p,t}^H, y_{1,t}^H, ... y_{e,t}^H, y_{1,t}^R, ... y_{j,t}^R]}^T,
\end{equation}
where, the upper script $(.)^H$ refers to the human pose and EMG signal, and $(.)^R$ refers to the robot joint angle configuration. 
The weight vector $\bm{\omega}$ for each demonstration is the concatenation of all weight vectors involved in the demonstration. Thus, all the interaction dimensions involved in the task are correlated as:
\begin{equation}
  \bm{\omega}_i = {[{(\bm{\omega}_1^H)}^T,...,{(\bm{\omega}_p^H)}^T,{(\bm{\omega}_1^H)}^T,...,{(\bm{\omega}_e^H})^T,{(\bm{\omega}_1^R)}^T...,{(\bm{\omega}_j^R)}^T]}^T.
\end{equation}
And, as in the single dimensional case, the weight vector is given as a linear regression model:
\begin{equation}
	p(\bm{y}_t|\bm{\omega}) = \mathcal{N}(\bm{y_t}| \bm{H}_t^T \bm{\omega}, 		{\bm{\Sigma}_y}),
\end{equation}
where, the $\bm{H}_t = diag({(\bm{\psi}_t^T)}_1,...,{(\bm{\psi}_t^T)}_{(p+e)},{(\bm{\psi}_t^T)}_1,...,{(\bm{\psi}_t^T)}_j)$ is the time-dependent basis matrix for the positions. 

Given the (partial) observations, we can compute the posterior distribution of both human and robot trajectories using a Kalman Filter. Where observations only contain human motion, thus robot observations are set to zero yielding: 
\begin{equation}
	\bm{y}_t^o = {[\bm{y}_{1,t}^H, ... \bm{y}_{p,t}^H, \bm{y}_{1,t}^H, ... 	\bm{y}_{e,t}^H, \bm{y}_{1,t}^R, ... \bm{y}_{j,t}^R]}^T.
\end{equation}
To contrast with a complete observation sequence $[t:t']$, the notation $[t-t'] \in {\mathbb{R}}^{s \times (p+e)}$ is used to indicate a sequence $s$ of partial observations in the interval (some measurements in the interval are missing). Observations can be considered as modulations to via-points. The operation is done by conditioning the ProMPs to reach a certain state $\bm{y}_{t-t'}^o$ at time $(t-t')$. The conditioning adds a desired observation to $\bm{x}_{t-t'}=[\bm{y}_{t-t'}^o, \bm{\Sigma}_y^o]$ 
to the probabilistic model and applying Bayes theorem. Kalman filtering is used to compute the posterior distribution as:
\begin{equation}
	\begin{aligned}
		\bm{\mu}_\omega^{new} &= \bm{\mu}_\omega + \bm{K} (\bm{y}_{t-t'}^o - \bm{H}_{t-t'} \bm{\mu}_\omega), \\
		\bm{\Sigma}_\omega^{new} &= \bm{\Sigma}_\omega - \bm{K} (\bm{H}_{t-t'} \bm{\Sigma}_\omega).
	\end{aligned}
\end{equation}
Here, $\bm{K}=\bm{\Sigma}_\omega \bm{H}_{t-t'}^T {(\bm{\Sigma}_y^o + \bm{H}_{t-t'} \bm{\Sigma}_\omega \bm{H}_{t-t'}^T)}^{-1}$. And, since missing observations exist, for each time step of the observation matrix $\bm{H}_{t-t'}$, the latter is set as:
\begin{equation}
  \bm{H}_{t-t'}
  =\begin{bmatrix}
  {(\bm{\psi}_t^T)}_1    & \cdots & 0 & 0 & \cdots & 0	\\
  0 					 & \ddots & 0 & 0 & \ddots & 0					\\
  0 & \cdots & {(\bm{\psi}_t^T)}_{(p+e)} & 0 & \cdots & 0\\
  0 & \cdots & 0 & 0_1 & \cdots & 0					\\
  0 & \ddots & \vdots & 0 & \ddots & 0 				\\
  0 & \cdots & 0 & 0 & \cdots & 0_j					\\
\end{bmatrix}
\end{equation}
with $\bm{H}_{t-t'} \in  {\mathbb{R}}^{(p+e+j)  \times (p+e+j)N}$.
\subsection{Phase estimation}
It's natural for a human to execute repetitions of a specific task with different speeds. The latter leads to uncertainty in the duration of the demonstration. To capture such spatial variation correctly, time alignment must been done. What's more, phase (or progress) analysis of human observations during testing must be estimated to aligning them to the trained spatial models. In our work, each demonstration was resampled yielding a nominal duration $T_{norm}$. As in \cite{2017IJRR-Maeda-PhaseEstimation}, we assume that the i$^{th}$ demonstration also has a constant temporal change in relation to the nominal duration and can define a scaling factor in Eqtn. \ref{eqtn:scaling_factor} to index all demonstrations by the nominal time index.
\begin{equation}
	\alpha_i = T_i / T_{norm}.
    \label{eqtn:scaling_factor}
\end{equation}
For phase estimation in testing, Maeda's single phase temporal model is used. And a distribution over phase rations from different demonstrations are modeled according to a normal distribution and set as the phase prior. We assume $\alpha \sim \mathcal{N}(\mu_\alpha, \sigma_\alpha)$. In testing, given a human observation $y_{t-t'}^o$, the posterior for the phase is computed as:
\begin{equation}
p(\alpha | \bm{y}_{t-t'}^o, \bm{\theta}) \propto 
p(\bm{y}_{t-t'} | \alpha, \bm{\theta}) p(\alpha), 
\end{equation}
where the $p(\alpha)$ is the prior probability of the scaling factor $\alpha$ as previously discussed. Additionally, the likelihood for a specific task is given as:
\begin{equation}
	p(\bm{y}_{t-t'} | \alpha, \bm{\theta}) = 
	\int p(\bm{y}_{t-t'}^o | \bm{\omega}, \alpha) p(\omega) d \bm{\omega}.
\end{equation}
For one specific task, given the human observations $y_{t-t'}^o$ the most probable scaling factor is:
\begin{equation}
	\alpha^* = \mathop{\arg\max}_{\alpha} p(\alpha | \bm{y}_{t-t'}^o, \bm{\theta})
\end{equation}
The best fit scaling factor $\alpha_k^*$ for each task is selected to get the set $\{ \alpha_k^*, \bm{\theta}_k \}$. Then, task recognition is done based on this set. 

\subsection{Task Recognition}\label{subsec:task_recognition}
We model a set of $k$ demonstrations from a probabilistic perspective and compute the posterior distribution of a task given human signal observations according to Eqtn. \ref{eqtn:task_recognition}
\begin{equation}
p(k | \bm{y}_{t-t'}^o) \propto p(\bm{y}_{t-t'}^o | \bm{\theta}_k, \alpha^*) p(k),
\label{eqtn:task_recognition}
\end{equation}

where, $p(k)$ is the task's prior probability and can be determined by the specific circumstances of an experiment. The likelihood of each component given the model $\theta$ is:
\begin{equation}
	p(\bm{y}_{t-t'}^o; \bm{\theta}_k, \alpha^*) = 
	\int p(\bm{y}_{t-t'}^o | \bm{H}_{t-t'}^o \bm{\omega}, \bm{\Sigma}_y) p(\bm{\omega}; 	\bm{\theta}_k) d \bm{\omega}.
\end{equation}
A task is selected by choosing the posterior with the highest probability: 
\begin{equation}
	k^{*} = \mathop{\arg\max}_{k} p(k | \bm{y}_{t-t'}^o)
\end{equation}

\begin{figure*}
\centering
	\includegraphics[width=\linewidth]{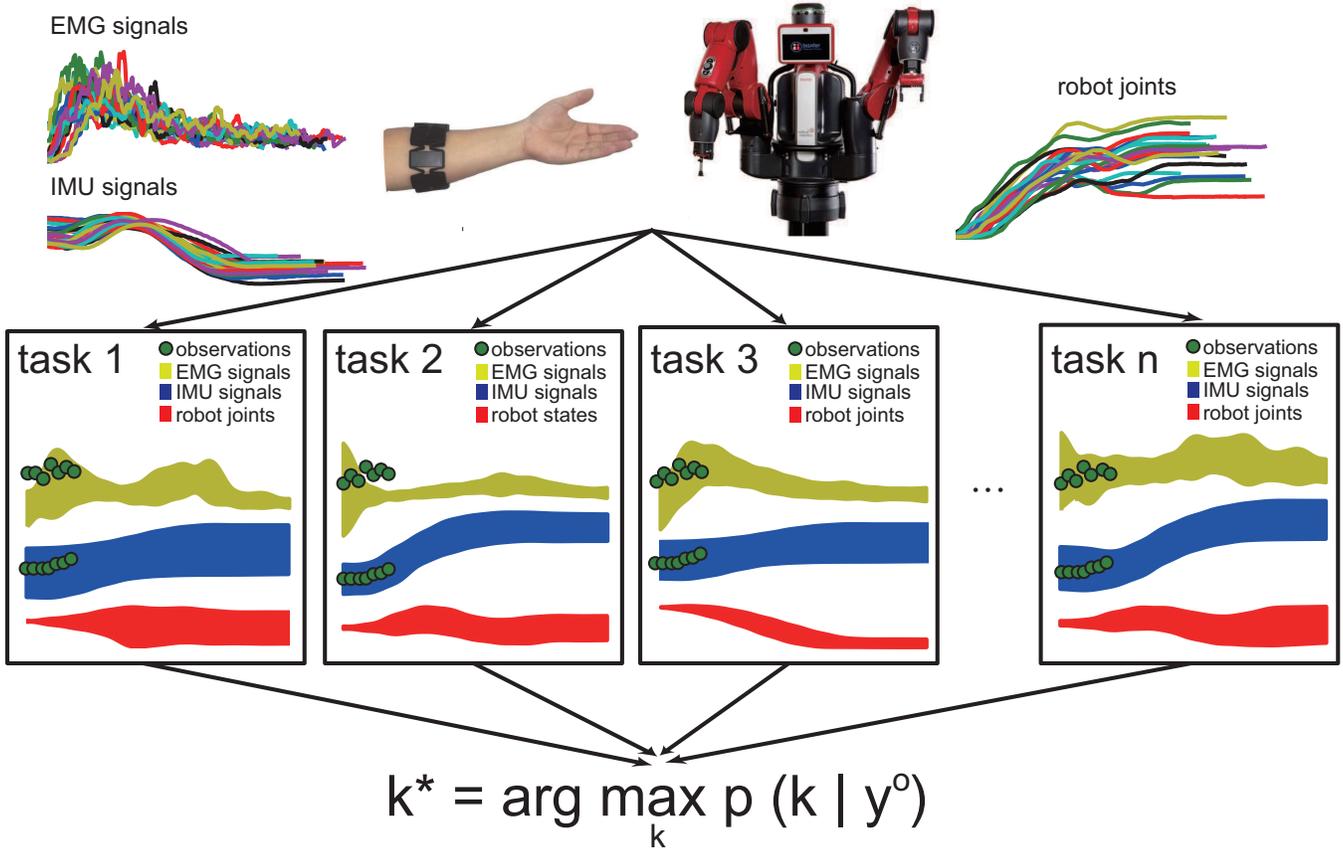}
 	\caption{Multidimensional Interaction ProMPs are used with an augmented state vector that integrates EMG-based muscle activity. Phase aligned trajectory way-points and EMG signals are parameterized into a lower dimensional weight space as a linear combination of basis functions. The correlation of all human-robot data is computed and the robot motion is inferred by computing a posterior probability distribution over the weights conditioned on the partial augmented human observation (shown in green circles). For task recognition, the task with the highest posterior probability of new observations given a task's probability is selected.}
	\label{fig:workflow}
\end{figure*}
Fig. \ref{fig:workflow}, summarizes the multiple task recognition problem. Motion and EMG signals from a human and robot states (joint angles or Cartesian) are captured. After demonstrating a collaborative task, we generate the probabilistic model for each task to represent multiple demonstrations using our method. For clarity sakes, sensor data is abstracted to a single dimension in the Figure. Note how human motion look similar across tasks. This condition leads to a situation where the likelihood for multiple tasks is very similar to each other, rendering it difficult to select a task with any certainty.
\section{Experiments and Results}\label{sec:experiments}
Our experimental testbed used a dual-armed upper-torso anthropomorphic Baxter robot, a Myo wearable armband and ROS Indigo in Linux Ubuntu 14.04. Kinesthetic teaching was used to drive Baxter in collaborative tasks. The Myo armband is composed of eight stainless steel EMG sensors and a nine axis IMU. Raw EMG and IMU data, along with motion, orientation, and rotation data are streamed over blue tooth. The band is placed around the forearm, as such it measures muscle signals in the forearms' anterior and posterior superficial muscles. Such data can play a vital complimentary role to motion data. 

To test the effects of EMG signals in Interaction ProMPs, three distinct hand-over tasks, but ones with similar human motions, were tested with and without EMG-signals. Namely, (i) passing an aluminum rod, (ii) passing a wrench, and (iii) passing a measuring tape. For the aluminum rod and measuring tape tasks, the human beings by grasping the corresponding object and then proceeds to pass them to the robot, the robot executes a parallel motion and picks the object. For the wrench task, it is the human who receives the tool from the Baxter robot. Each of the three tasks was repeated 10, 15, and 20 times respectively for training and an additional 10 trials for testing. The different number of training trials was set to study the impact of training trials with EMG signals. All trial data was time aligned by resampling. Fig \ref{fig:collab_task_example}, shows a snap shot for each of the three experiments at the time the tool is handed over.
For the three tasks the human motion is nearly the same: each experiment has the human standing in approximately the same location and the arm pose is also started in approximately the same location. Such assumptions are really realistic given that a work site has an established working environment. Similar motions are attempted by a single user each time. This sets the stage to measure the task recognition ability when using the EMG signals. We report results for experiments with orientation and orientation with EMG data (we did not in fact use a Cartesian trajectory due to the noisiness of IMU motion data). Both under different amounts of human observations: 10\% and 20\% of the duration of the task.

\subsection{Results}\label{subsec:results}
We present results in a set of tables. Each table presents the results according to the number of training demonstrations as well as the human observation ratio for the task, and the recognition accuracy result for the three tasks with and without EMG data. Table \ref{tbl:20_0.1} shows results for 20 demonstrations and 10\% observation ratio. Table \ref{tbl:15_0.1}: 15 and 10\% respectively, Table \ref{tbl:10_0.1} 10 and 10\% respectively, and Table \ref{tbl:10_0.2} 10 and 20\% respectively.
\begin{table}[h!]
	\centering
	\caption{number of demonstration: 20, observation ratio: 0.1}
	\begin{tabular}{l|cc}
	\textit{Task} & w/out EMG & with EMG\\
	\midrule
	  Aluminum Rod					& 0.90 			& \red{1.00} 	\\
      Wrench 			   			& 0.60 			& \red{1.00} 	\\
      Measuring Tape	           	& 0.10 			& \red{0.70} 	\\
	\label{tbl:20_0.1}
    \end{tabular}
\end{table}    
\begin{table}[h!]
	\centering
	\caption{number of demonstration: 15, observation ratio: 0.1}
	\begin{tabular}{l|cc}
	\textit{Task} & without EMG & with EMG\\
	\midrule
	  Aluminum Rod					& 0.60 			& \red{1.00} 	\\
      Wrench			   			& 0.60 			& \red{0.90} 	\\
      Measuring Tape   	            & 0.10 			& \red{0.70} 	\\
	\label{tbl:15_0.1}
    \end{tabular}
\end{table}    
\begin{table}[h!]
	\centering
	\caption{number of demonstration: 10, observation ratio: 0.1}
	\begin{tabular}{l|cc}
	\textit{Task} & without EMG & with EMG\\
	\midrule
	  Aluminum Rod 					& 0.00 			& \red{0.50} 	\\
      Wrench 			   			& 0.00 			& \red{0.80} 	\\
      Measuring Tape	           	& \red{0.80} 	& 0.70 			\\
	\label{tbl:10_0.1}
    \end{tabular}
\end{table}    
\begin{table}[h!]
	\centering
	\caption{number of demonstration: 10, observation ratio: 0.2}
	\begin{tabular}{l|cc}
	\textit{Task} & without EMG & with EMG\\
	\midrule
	  Aluminum Rod					& 0.30 			& \red{1.00} 	\\
      Wrench			   			& 1.00 			& \red{1.00} 	\\
      Measuring Tape	           	& 0.90 			& \red{1.00} 	\\
	\label{tbl:10_0.2}
    \end{tabular}
\end{table}    

We note that out of the 12 measurements that we made (different number of demonstrations \& observation ratios by the three tasks), 11 out of the 12 tasks or 91.6\%, experiments with EMG signals out-classified those without. Not only so, if we average classification rates across all experiments, we see that without EMG signals we had an accuracy of get a sum of 49.2\%, while for the augmented EMG signals we get an accuracy of 85.8\%. That is 74.6\% more accurate recognition (see our supplemental information for details \cite{2017Humanoids-Chen-supplementalURL}). In summary, integration of EMG signals not only is correct more than without, but is also does it overwhelmingly better. We believe this too would have significant effects in user-load (mental, emotional, and physical) as the robot would handle tasks in ways that do not require the human to adjust its handling, thus enhancing the overall experience. However, this is was not formally measured in this study.

We noted that during task recognition inference, there is a strong dependence on the prior. That is, observations make a small contribution. For motion trajectory only demos, failed task recognition predictions result in wrong robot collaborative motions. But with the integration of EMG-based muscular signals to human motion observations, the distinct EMG signatures disambiguate task recognition yielding large probabilistic differences across tasks.
\section{Discussion}\label{sec:discussion}
Our work demonstrates that the integration of EMG-based muscular activity into Interaction ProMPs for tasks with similar motions significantly increased task recognition discernment. It was shown that for three different hand-over tasks (including human-to-robot and robot-to-human passes) with different number of training demonstrations and different number of human observation ratios, experiments with EMG-signals overwhelmingly outperformed those without, that is by 74.6\%. 

This result shows that human muscular activity can significantly augment a robot's insight into human service tasks and improve its task recognition. This in turn allows a robot to improve how it handles an object: it's end-pose at the time of the hand-over and possibly its dynamics. In doing so, hand-overs and numerous other tasks would place a lower user-load on the human: mentally, physically, and emotional. If the robot does not need to adjust his own pose upon a handover because the robot has correctly reached an object and thereafter properly handled, the human would be at greater ease. We leave it to future work to show the quantitative effects of this work. While the proposed methodology of our work did not differ from that in \cite{2017IJRR-Maeda-PhaseEstimation}, we believe that the knowledge and insight gained from our analysis of a rarely used biometric signal in HRC offers a relevant insight to the field. We estimated this may be the first work that studies the impact of muscular activity in human robot collaboration tasks. 

There are a number of enhancements we set as future work. First, is to explore more compelling cases for the use of muscular-based EMG-signals in physical human interaction. The authors believe that a wide array of possibilities can exist through haptic communication with the robot. That is, through direct physical touch. EMG can serve as a primary signal, especially if finger motion cannot be tracked or visual occlusion prevents identifying small nuanced haptic motions. Other improvements to the current work include the use of non-parametric methods to estimate an optimal number of basis functions in modeling trajectories. This will result in better modeling, particularly when tasks have more complex dynamics. Bayesian estimation can also yield more confident beliefs in computing relevant parameters as opposed to MAP estimates. 

\section{Conclusion}\label{sec:conclusion}
We proposed the integration of EMG-based muscular activity into the Interaction ProMP framework to augment the human observation representation. A probabilistic model containing the variance of human and robot motion and (forearm) muscle activity was used. Motion Primitive's temporal distribution were modeled through a Hierarchical Bayesian Model with Gaussian distributions. A temporal sequence distribution is obtained from demonstrations and the correlation across all dimensions jointly modeled and used to generate a corresponding robot motion from the observation of human action signals. The result was an increased capacity to discern tasks with similar trajectories but different tools aiding the robot to improve object handling and reducing user-load. 
\section{Acknowledgements} \label{sec:Acknowledgements}
This work is supported by ``Major Project of the Guangdong Province Department for Science and Technology (2014B090919002), (2016B0911006).''
\bibliographystyle{IEEEtran}
\bibliography{IEEEabrv,Xbib}

\begin{thebibliography}{10}
\providecommand{\url}[1]{#1}
\csname url@samestyle\endcsname
\providecommand{\newblock}{\relax}
\providecommand{\bibinfo}[2]{#2}
\providecommand{\BIBentrySTDinterwordspacing}{\spaceskip=0pt\relax}
\providecommand{\BIBentryALTinterwordstretchfactor}{4}
\providecommand{\BIBentryALTinterwordspacing}{\spaceskip=\fontdimen2\font plus
\BIBentryALTinterwordstretchfactor\fontdimen3\font minus
  \fontdimen4\font\relax}
\providecommand{\BIBforeignlanguage}[2]{{%
\expandafter\ifx\csname l@#1\endcsname\relax
\typeout{** WARNING: IEEEtran.bst: No hyphenation pattern has been}%
\typeout{** loaded for the language `#1'. Using the pattern for}%
\typeout{** the default language instead.}%
\else
\language=\csname l@#1\endcsname
\fi
#2}}
\providecommand{\BIBdecl}{\relax}
\BIBdecl

\bibitem{2017Humanoids-Chen-supplementalURL}
\BIBentryALTinterwordspacing
L.~Chen, ``Supplement to learning human-robot collaboration insights through
  the integration of muscular activity in interaction motion models,'' 2017.
  [Online]. Available: \url{http://www.juanrojas.net/ipromp_emg/}
\BIBentrySTDinterwordspacing

\bibitem{Amor2016Interaction}
H.~B. Amor, G.~Neumann, S.~Kamthe, and O.~Kroemer, ``Interaction primitives for
  human-robot cooperation tasks,'' in \emph{IEEE International Conference on
  Robotics and Automation}, 2016, pp. 2831--2837.

\bibitem{Schaal2006Dynamic}
S.~Schaal, \emph{Dynamic Movement Primitives -A Framework for Motor Control in
  Humans and Humanoid Robotics}.\hskip 1em plus 0.5em minus 0.4em\relax
  Springer Tokyo, 2006.

\bibitem{Ijspeert2013Dynamical}
A.~J. Ijspeert, J.~Nakanishi, H.~Hoffmann, P.~Pastor, and S.~Schaal,
  ``Dynamical movement primitives: learning attractor models for motor
  behaviors.'' \emph{Neural Computation}, vol.~25, no.~2, pp. 328--373, 2013.

\bibitem{Vogt2015Behavior}
D.~Vogt, B.~Lorenz, S.~Grehl, and B.~Jung, ``Behavior generation for
  interactive virtual humans using context‐dependent interaction meshes and
  automated constraint extraction,'' \emph{Computer Animation \& Virtual
  Worlds}, vol.~26, no. 3-4, pp. 227--235, 2015.

\bibitem{Vogt2017A}
D.~Vogt, S.~Stepputtis, S.~Grehl, B.~Jung, and H.~B. Amor, ``A system for
  learning continuous human-robot interactions from human-human
  demonstrations,'' in \emph{IEEE International Conference on Robotics and
  Automation}, 2017.

\bibitem{Vogt2017Learning}
D.~Vogt, S.~Stepputtis, R.~Weinhold, B.~Jung, and H.~B. Amor, ``Learning
  human-robot interactions from human-human demonstrations (with applications
  in lego rocket assembly),'' in \emph{Ieee-Ras International Conference on
  Humanoid Robots}, 2017.

\bibitem{2017IJRR-Maeda-PhaseEstimation}
G.~Maeda, M.~Ewerton, G.~Neumann, R.~Lioutikov, and J.~Peters, ``Phase
  estimation for fast action recognition and trajectory generation in
  human--robot collaboration,'' \emph{The International Journal of Robotics
  Research}, p. 0278364917693927, 2017.

\bibitem{2015ICRA-ewerton-LearnMultCollabTasks_MixtureInteractionPrimitives}
M.~Ewerton, G.~Neumann, R.~Lioutikov, H.~B. Amor, J.~Peters, and G.~Maeda,
  ``Learning multiple collaborative tasks with a mixture of interaction
  primitives,'' in \emph{Robotics and Automation (ICRA), 2015 IEEE
  International Conference on}.\hskip 1em plus 0.5em minus 0.4em\relax IEEE,
  2015, pp. 1535--1542.

\bibitem{Maeda2016Probabilistic}
G.~J. Maeda, G.~Neumann, M.~Ewerton, R.~Lioutikov, O.~Kroemer, and J.~Peters,
  ``Probabilistic movement primitives for coordination of multiple
  human–robot collaborative tasks,'' \emph{Autonomous Robots}, pp. 1--20,
  2016.

\bibitem{Calinon2004Gesture}
S.~Calinon and A.~Billard, ``Gesture recognition and reproduction for a
  humanoid robot using hidden markov models,'' in \emph{Ami/pascal/im2/m4
  Workshop on Multimodal Interaction and Related Machine Learning Algorithms},
  2004.

\bibitem{calinon2006learning}
S.~Calinon, F.~Guenter, and A.~Billard, ``On learning the statistical
  representation of a task and generalizing it to various contexts,'' in
  \emph{Robotics and Automation, 2006. ICRA 2006. Proceedings 2006 IEEE
  International Conference on}.\hskip 1em plus 0.5em minus 0.4em\relax IEEE,
  2006, pp. 2978--2983.

\bibitem{Ho2010Spatial}
E.~S.~L. Ho, T.~Komura, and C.~L. Tai, ``Spatial relationship preserving
  character motion adaptation,'' in \emph{ACM SIGGRAPH}, 2010, p.~33.

\bibitem{Paraschos2013Probabilistic}
A.~Paraschos, C.~Daniel, J.~Peters, and G.~Neumann, ``Probabilistic movement
  primitives,'' \emph{Advances in Neural Information Processing Systems}, pp.
  2616--2624, 2013.

\bibitem{Koert2017Demonstration}
D.~Koert, G.~Maeda, R.~Lioutikov, G.~Neumann, and J.~Peters, ``Demonstration
  based trajectory optimization for generalizable robot motions,'' in
  \emph{Ieee-Ras International Conference on Humanoid Robots}, 2017, pp.
  515--522.

\bibitem{2008TroHap-Reed-PhysicalCollab-HH_HR}
K.~B. Reed and M.~A. Peshkin, ``Physical collaboration of human-human and
  human-robot teams,'' \emph{IEEE Transactions on Haptics}, vol.~1, no.~2, pp.
  108--120, 2008.

\bibitem{Peternel2016Adaptation}
L.~Peternel, N.~Tsagarakis, D.~Caldwell, and A.~Ajoudani, ``Adaptation of robot
  physical behaviour to human fatigue in human-robot co-manipulation,'' in
  \emph{IEEE RAS International conference on humanoid robotics}, 2016.

\bibitem{2007Tro-kulic-AffectiveStateEstimationHRI}
D.~Kulic and E.~A. Croft, ``Affective state estimation for human--robot
  interaction,'' \emph{IEEE Transactions on Robotics}, vol.~23, no.~5, pp.
  991--1000, 2007.

\end{thebibliography}

\end{document}